\documentclass[runningheads]{llncs}
\usepackage[T1]{fontenc}
\usepackage{graphicx}
\usepackage{biblatex}
\usepackage{caption}
\usepackage{subcaption}
\usepackage[hidelinks]{hyperref}
\usepackage{xcolor}
\def\BibTeX{{\rm B\kern-.05em{\sc i\kern-.025em b}\kern-.08em
    T\kern-.1667em\lower.7ex\hbox{E}\kern-.125emX}}
\bibliography{references}

\newcommand{\modtext}[1]{\textcolor{black}{#1}}

\begin{document}

\title{The Impact of Robots' Facial Emotional Expressions on Light Physical Exercises\thanks{We acknowledge the support of the Natural Sciences and Engineering Research Council of Canada (NSERC), funding reference number RGPIN-2022-03857.  \textit{This is a prerint version of the paper presented at the International Conference on Social Robotics (ICSR) 2023.}}}
\titlerunning{The Impact of Robots' Facial Expressions on Light Physical Exercises}
\author{Nourhan Abdulazeem\orcidID{0009-0003-3706-0103} \and
Yue Hu\orcidID{0000-0002-3846-9096}}
\authorrunning{N. Abdulazeem and Y. Hu}
\institute{Active \& Interactive Robotics Lab, Department of Mechanical and Mechatronics Engineering, University of Waterloo, N2L3G1, Waterloo, Ontario, Canada \\
\email{\{nourhan.abdulazeem, yue.hu\}@uwaterloo.ca}}
\maketitle              % typeset the header of the contribution
\begin{abstract}
To address the global challenge of population aging, our goal is to enhance successful aging through the introduction of robots capable of assisting in daily physical activities and promoting light exercises, which would enhance the cognitive and physical well-being of older adults. Previous studies have shown that facial expressions can increase engagement when interacting with robots. This study aims to investigate how older adults perceive and interact with a robot capable of displaying facial emotions while performing a physical exercise task together. We employed a collaborative robotic arm with a flat panel screen to encourage physical exercise across three different facial emotion conditions. We ran the experiment with older adults aged between 66 and 88. Our findings suggest that individuals perceive robots exhibiting facial expressions as less competent than those without such expressions. Additionally, the presence of facial expressions does not appear to significantly impact participants' levels of engagement, unlike other state-of-the-art studies. This observation is likely linked to our study's emphasis on collaborative physical human-robot interaction (pHRI) applications, as opposed to socially oriented pHRI applications. Additionally, we foresee a requirement for more suitable non-verbal social behavior to effectively enhance participants' engagement levels.
% Current abstract is 166 words. Findings and take-home message are still not added.
% The abstract should briefly summarize the contents of the paper in 150--250 words.

\keywords{facial expressions  \and emotions \and  collaborative robotic arms \and social-physical human-robot interaction}
\end{abstract}

% facial expressions, emotions,  collaborative robotic arms, social-physical human-robot interaction

\section{Introduction}
\label{introduction}
For the first time in history, the global population of individuals aged 6 and above is projected to exceed the number of younger people. The decline in fertility rates and the increase in life expectancy have resulted in a global phenomenon of population aging \cite{noauthor_ageing_nodate}. This attracts researchers' attention to studying how to enhance older adults' life quality and independent living.

A potential enhancement for successful aging is introducing robots that can provide physical assistance with essential daily activities and promote light physical exercises. It has been proven that light physical exercise can be beneficial for older adults to maintain their cognitive and physical well-being \cite{fitter_exercising_2020}. As a consequence, this has drawn our attention to exploring what could contribute to a successful physical human-robot interaction (pHRI) \cite{losey_review_2018}.

Recent studies have demonstrated that, by bridging social human-robot interaction (sHRI) and pHRI, robots can physically interact with humans while also being socially acceptable \cite{fitter_exercising_2020}. One approach that has proven its effectiveness in enhancing robot perception and engagement is endowing robots with the ability to exhibit facial emotional expressions \cite{urakami_nonverbal_2022}. This has motivated us to explore the potential of facial emotional expressions in pHRI scenarios.

\begin{figure}
     \centering
     \begin{subfigure}[b]{0.48\textwidth}
         \centering
         \includegraphics[width=0.9\textwidth]{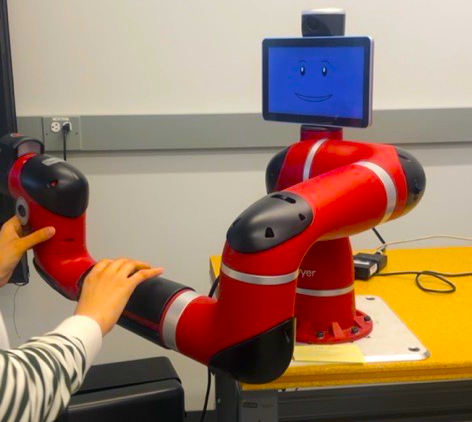}
         \caption{Before performing a push and robot is in initial configuration}
         \label{Sawyer_initial_config}
     \end{subfigure}
     \hfill
     \begin{subfigure}[b]{0.48\textwidth}
         \centering
         \includegraphics[width=0.9\textwidth]{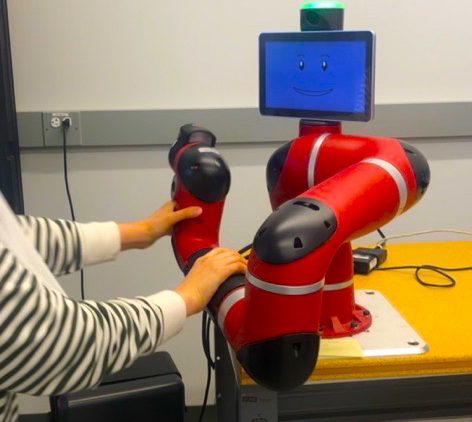}
         \caption{After a push is performed and robot headlight blinks}
         \label{Sawyer_counted_push}
     \end{subfigure}
        \caption{Participant exercising with Sawyer in the unresponsive social behavior condition}
        \label{fig:physical_exercise}
\end{figure}

In this study, we adopt a light physical exercise scenario, as one of the potential applications for physically interactive robots in domestic environments, to investigate the impact of facial emotional expressions on users' perception of the robot and their level of engagement. We utilized Sawyer, a collaborative robotic arm developed by Rethink Robotics, for this purpose. Sawyer is equipped with a flat panel screen that allows us to display various facial expressions, as shown in Figure \ref{fig:physical_exercise}. We anticipate that our results will be helpful for the research community toward the exploration of effective  social skills for physically interactive robots.

% If your review is accepted for publishing, replace all the citations in the following paragraph with it.
Our research is founded upon two primary domains: sHRI and pHRI, which serve as the pillars of our investigation. Remarkable efforts were made to explore social-physical robots in various contexts such as hugging \cite{block_arms_2022}, touching in social and nursing scenarios \cite{mazursky_physical_2022}, handshaking \cite{law_touching_2021}, and playing games \cite{fitter_how_2020}. However, only limited efforts involved the investigation of both domains with robots that possess high dexterity and manipulation capabilities \cite{abdulazeem_human_2023} which are essential qualities for robots to efficiently engage in physical interactions. Even fewer studies have devoted their efforts to investigating facial emotional expressions, for those types of robots \cite{fitter_how_2020}.

Some existing studies have relied on facial expressions to enhance the user's engagement with robots that possess high dexterity and manipulation capabilities in various pHRI scenarios, such as physical exercise \cite{fitter_exercising_2020}, clapping/gaming, and teaching \cite{abdulazeem_human_2023}. However, they did not investigate the impact of the robot's facial expressions on the interaction, unlike Tsalamlal et al. \cite{tsalamlal_affective_2015} and Fitter et al. \cite{fitter_how_2020}. Tsalamlal et al. \cite{tsalamlal_affective_2015} investigated how participants combine facial expressions and handshakes to assess the perceived emotions in robots. The findings indicated that participants assigned greater significance to facial expressions when evaluating Valence. Fitter et al. \cite{fitter_how_2020} assessed how participants' emotions were affected by a robot's responsive facial expressions compared to an unresponsive robot's facial expressions during a hand-clapping game. Participants perceived the interactive face as more pleasant, energetic, and less robotic than the unresponsive one.

However, none of the previously mentioned studies have investigated participants' perceptions of the robot's characteristics, such as its perceived intelligence. Furthermore, the influence of these facial expressions on users' performance remains unexplored. To address these gaps, our study seeks to investigate participants' perceptions of a physically interacting robot displaying facial expressions and examine the potential impact of these expressions on their performance during physical interactions.

\section{Research Questions}
\label{research_questions}
Our study is designed to build upon the insights gained from existing literature. By rigorously exploring the effects of facial emotional expressions on older adults engaged in pHRI applications. As a result, the following research questions were formulated to lead this study:
\begin{itemize}
    \item \textbf{RQ1}: Will facial emotional expressions impact an older adult's level of engagement and perception of  a robot in a pHRI scenario? 
\end{itemize}
The answer to the first research question will help us understand the importance of relying on facial  expressions as a means of communication during pHRI with older adults. These findings will guide the research community towards investing further efforts in the development of facial emotional expressions for successful pHRI. Alternatively, they may prompt exploration of other social behaviors that could be better suited for typical pHRI scenarios.

To comprehend the impact of facial expression responsiveness, as well as the mere presence of a robot's face regardless of its responsiveness, on participants' perceptions of the robot's characteristics and performance, we pose our second research question:
\begin{itemize}
    \item \textbf{RQ2:} Does a responsive robot's facial emotional expressions impact an older adult's level of engagement and perception of a robot, compared to an unresponsive robot, in a pHRI scenario?
\end{itemize}

\section{Method}
\label{method}
\subsection{Settings}
\label{settings}
To answer the proposed research questions, a user study,  in the form of a light physical exercise game, was conducted in the Active \& Interactive Laboratory at the University of Waterloo. The objective of the exercise is to perform the highest number of pushes possible, according to each participant's comfortable pace, against the robot's 4th joint as shown in Figure \ref{fig:physical_exercise}. Further details of the game design are provided in Section \ref{game_design}.

A between-subject study design is considered and Sawyer from Rethink Robotics is used. Sawyer is a 7-degree-of-freedom torque-controlled manipulator equipped with a flat-panel screen and headlight, as shown in Figure \ref{fig:physical_exercise}.

\begin{figure}
     \centering
     \begin{subfigure}[b]{0.3\textwidth}
         \centering
         \includegraphics[width=\textwidth]{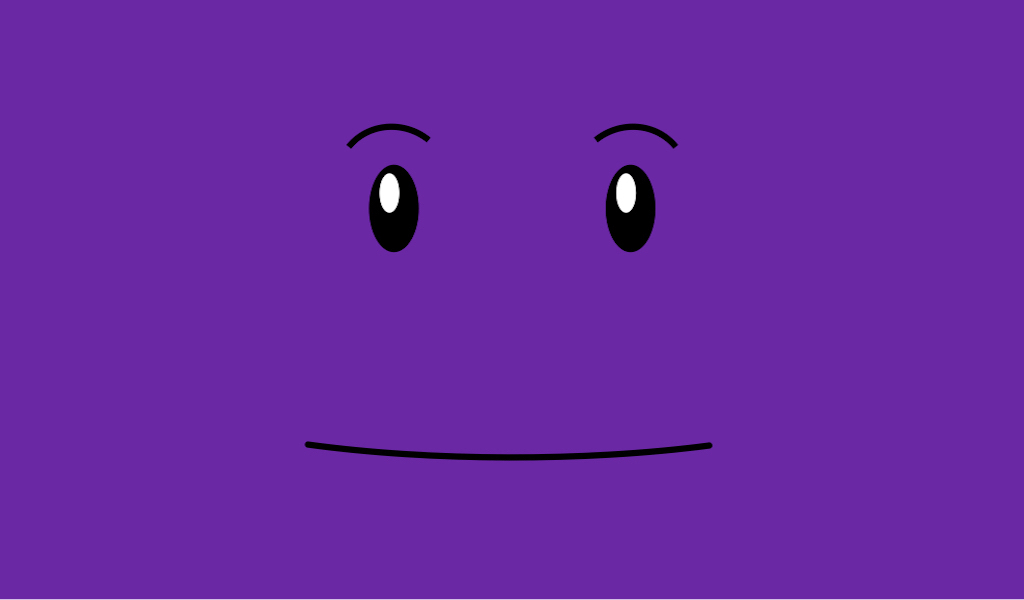}
         \caption{Neutral Face}
         \label{Neutral_Face}
     \end{subfigure}
     \hfill
     \begin{subfigure}[b]{0.3\textwidth}
         \centering
         \includegraphics[width=\textwidth]{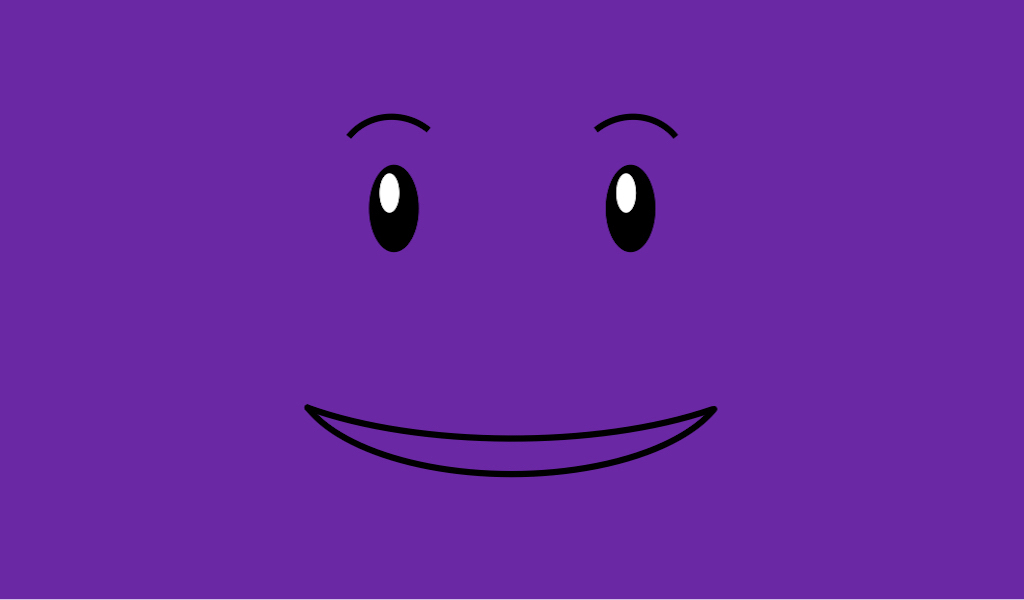}
         \caption{Happy Face}
         \label{Happy_Face}
     \end{subfigure}
     \hfill
     \begin{subfigure}[b]{0.3\textwidth}
         \centering
         \includegraphics[width=\textwidth]{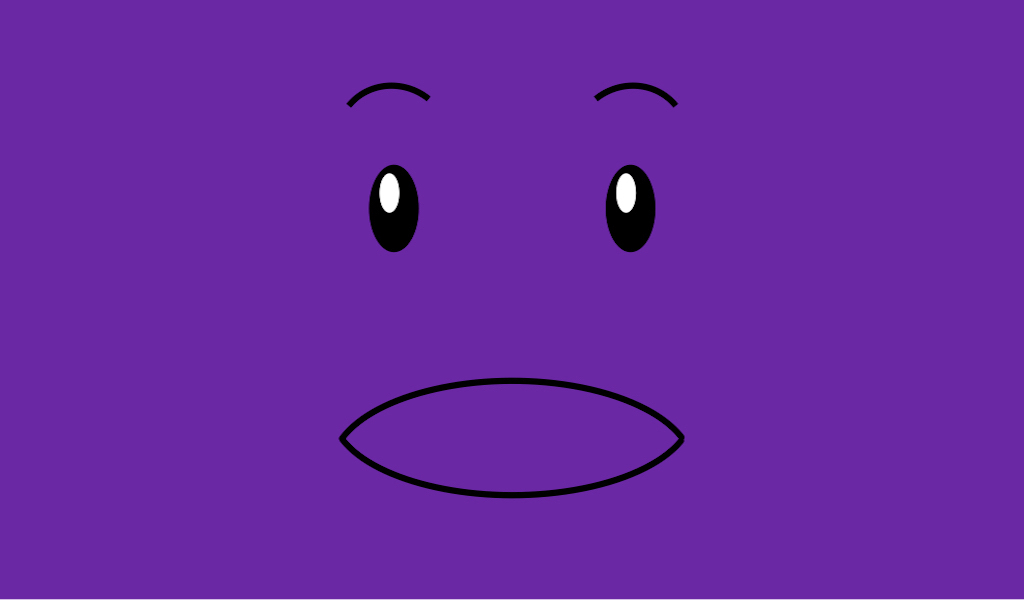}
         \caption{Surprised Face }
         \label{Surprised_Face}
     \end{subfigure}
        \caption{Sawyer facial emotional expressions \cite{fitter_designing_2016}}
        \label{Sawyer_facial_emotional_expressions}
\end{figure}

During participants exercising with the robot, 3 conditions of facial emotional expressions were considered:
\begin{enumerate}
    \item \textbf{Inactive Facial Expression}: The robot displays its default screen, which features the Rethink Robotics logo \cite{rethink_robotics_rethinkrobotics_nodate}.
    \item  \textbf{Unresponsive Facial Expression}: The robot showcases a happy face (as depicted in Figure \ref{Happy_Face}).
    \item \textbf{Responsive Facial Expression}: The robot exhibits varying facial emotional expressions in response to the user's performance.
\end{enumerate}

In this paper, we will also refer to the unresponsive and responsive conditions together as the active conditions. In the responsive condition, the robot shows a neutral face (shown in Figure \ref{Neutral_Face}) at the beginning of the interaction. After the participant performs a set of successful pushes, the robot shows a happy face (shown in Figure \ref{Happy_Face}) and after another set of successful pushes, the robot shows a surprised face (shown in Figure \ref{Surprised_Face}).

We opted to utilize Fitter and Kuchenbecker's established facial emotional expression set \cite{fitter_designing_2016} due to its cross-cultural evaluation, a critical factor for conducting experiments in a Canadian societal context, as in our case. \modtext{Fitter et al. \cite{fitter_designing_2016} found that participants from the USA and India, similar to our participants (Caucasians and Southeast Asians), successfully identified their proposed set of the facial emotional expressions.} Moreover, the study, which took place online, exclusively presented participants with Baxter's head. It's worth noting that Baxter's head is almost identical to Sawyer's head, as both robots are products of Rethink Robotics. 

We considered using the safest rated facial emotions, according to Fitter and Kuchenbecker's results, as safety is the most crucial human factor in HRI \cite{coronado_evaluating_2022}. Therefore, we decided to use the neutral, happy, and surprised faces for the responsive condition and the happy face, which is rated the safest among all faces, for the unresponsive condition. While the red and the purple colors were rated as the most energetic face colors, we chose to use purple for all conditions as red was, also, rated the least pleasant. Similarly, we aimed at using an arousing color as it contributes significantly to promoting the interaction. It should be noted that Sawyer's face color in Figure \ref{fig:physical_exercise} is purple, but it is shown blue due to the camera effect.

\subsubsection{Physical Exercise Design}
\label{game_design}
The exercise starts with Sawyer at its initial joint configuration, as shown in Figure \ref{Sawyer_initial_config}. The participants stood upright and faced Sawyer’s head. They are asked to perform the highest number of pushes possible according to their comfortable pace in 1 minute against Sawyer’s 4th joint as shown in Figure \ref{fig:physical_exercise}. Further details about the instructions provided for the participants are indicated in Section \ref{procedure}. A push is only counted if a participant was able to push the joint to make an angle offset greater than a pre-defined threshold. Each time a push is counted, Sawyer’s green headlight blinks as an indicator for the participant. Figure \ref{Sawyer_counted_push} shows the robot state, note the headlight, when a push is counted. Example of a participant exercising\footnote{\href{https://youtu.be/mZJcMLABNHg}{https://youtu.be/mZJcMLABNHg}}.

\subsubsection{Robot Control}
\label{robot_control}
In order to physically exercise with Sawyer in a fully-autonomous mode, it is required to be under joint impedance control. The joints' initial configuration and stiffness are kept constant across all conditions, whereas Sawyer's facial expressions are adjusted according to the condition being examined.

In the responsive condition, to keep a consistent interaction experience with Sawyer across all participants and account for differences in participants' physical capabilities, each participant performed a trial session to determine their capability of pushing Sawyer's joints, i.e., baseline. Further details on how the trial session is conducted are in Section \ref{procedure}. Thus, we were able to predict each participant's total number of pushes during the actual session. Accordingly, we implemented our code to show a happy face after 25\% of the baseline, and a surprised face after 75\% of the baseline. Hence, each participant in the responsive conditions gets to experience all the facial expression alterations around the same phase in the experiment despite the expected diversity in participants' physical capabilities.

\subsection{Participants}
\label{participants}
Twenty-seven participants were recruited in our study (17 female (F); 10 male (M), all older adults) from the University of Waterloo Research in Aging Participant Pool (WRAP), between the ages of 66 and 88 years (M = 76.52, SD = 6.12). Out of the 27 participants, all 3 conditions were randomly assigned 9 participants each (ages: M = 76.78, SD = 7.31, 5 F, 4 M for the inactive condition, ages: M = 77.44, SD = 4.27, 6 F, 3 M for the unresponsive condition, ages: M = 75.33, SD = 6.18, 6 F, 3 M for the responsive condition).

Among the recruited participants for the experiment, the majority were right-handed. However, in the inactive condition, there were four exceptions: two left-handed individuals and two who identified as ambidextrous. Additionally, each of the two active conditions included one left-handed participant. Ethnically, the majority of participants identified as Caucasian\modtext{,with two participants identifying as Southeast Asian.} Notably, two participants of South Asian descent were assigned to each of the active conditions. Furthermore, all participants demonstrated good eyesight as confirmed by a brief eye test.

Initial survey results indicated that all participants displayed normal levels of depression, stress, and anxiety according to the Depression Anxiety Stress Scale 21 (DASS-21) \cite{oei_using_2013}. Notably, depression levels were evaluated due to their established influence on activity motivation, a practice observed in prior similar experiments \cite{fitter_exercising_2020}. None of the participants had prior exposure to the Sawyer robot, as confirmed by a 5-point Likert scale. While some participants had encountered other robots before, over 50\% of the assigned participants had no previous robotics experience across all conditions. During the trial session, participants demonstrated closely matched physical capabilities. Additionally, none of the recruited participants had upper or lower limb motion disabilities.

All our experiments received ethical approval from the University of Waterloo Human Research Ethics Board (protocol N. 45340) at the University of Waterloo, Ontario, Canada. Before the experiment, participants received proper information and gave informed consent to participate in the study.

\subsection{Procedure}
\label{procedure}
Each participant visited the laboratory and dedicated 20 to 30 minutes to complete the study. After obtaining participants' informed consent, they provided demographic information including age, gender, ethnicity, profession, and handedness. Following this, participants watched an instructional video \footnote{\href{https://youtu.be/HxXZVLemShQ}{https://youtu.be/HxXZVLemShQ}} on how to exercise with the robot, without indicating the robot's capability of facial expressions, i.e., its screen is not shown. The decision to withhold information about the robot's facial expression capabilities prior to the experiment was intentional, aiming to prevent the formation of unrealistic expectations.

A short eye test was then administered to ensure participants' ability to see the robot's screen clearly. At the beginning of the setup, Sawyer's screen was turned away from the participant to prevent visibility of the screen. Participants were given the opportunity to perform 2 to 3 pushes to become accustomed to the robot's stiffness and determine their preferred distance from it. The experimenter ensured that participants performed the exercise correctly by ensuring they understood how the pushes were being counted. 

Next, a trial session lasting 10 seconds was conducted with Sawyer's screen still turned away. The purpose was to gauge each participant's physical capability. Upon successful completion of the trial, the robot rotated its screen to face the participant and proceeded to execute one of the three conditions detailed in Section \ref{settings}. Participants conducted the trial session with the robot in the same state as depicted in the instructional video (Sawyer's screen not facing the participant). Participants performed the actual session for a duration of 1 minute.

Following the task, participants completed robot perception and engagement questionnaires, detailed in Section \ref{measures}. Subsequently, a debriefing session was held to address any questions or concerns. As a token of appreciation for their time, each participant received remuneration.

\subsection{Measures}
\label{measures}

To evaluate robot perception, we employed the Robot Social Attribute Scale (RoSAS) \cite{carpinella_robotic_2017}. RoSAS measured participants' perceived competence, warmth, and discomfort \modtext{on a 9-point scale}. Additionally, perceived safety was evaluated using the corresponding subscale from the Godspeed questionnaire \cite{bartneck_measurement_2009} \modtext{, which employs a 5-point scale}. Perceived trust was assessed with a single-item questionnaire employing a 5-point Likert Scale. Participants indicated their level of trust by responding to the statement 'I trust the robot,' where 1 denoted 'strongly agree' and 5 denoted 'strongly disagree'. Items within each sub-scale are randomized.

For engagement assessment, we employed both objective and subjective evaluation methods. Participants indicated their level of engagement by responding to the statement 'I felt engaged with the robot during exercising' using a 5-point Likert Scale, where 1 represented 'strongly agree' and 5 represented 'strongly disagree'. The objective assessment \((E_{obj})\) involved determining the ratio between the actual number of pushes during the session \((P_{actual})\) and the expected number of pushes \((P_{expected})\) calculated from the trial session.
\[E_{obj} = \frac {P_{actual}} {P_{expected}} \]
Furthermore, we sought to understand the reasons behind the participants' responses to our quantitative measures by relying on open-ended questions. These questions included: "Did you enjoy the exercising session? Why or why not?", "What do you think about Sawyer as an exercising partner?", "What stood out to you the most about interacting with the robot?", "Do you think the robot can have more features that would make it more interesting? Suggest features.", and for the responsive condition, "Did you observe any changes in the robot's facial expressions? If so, how would you describe the changes in its facial expressions?"   To ensure thorough and meaningful responses, we specifically asked participants to provide the reasoning behind their answers.

\begin{figure}[!h]
    \centering
    \includegraphics[width=1\linewidth]{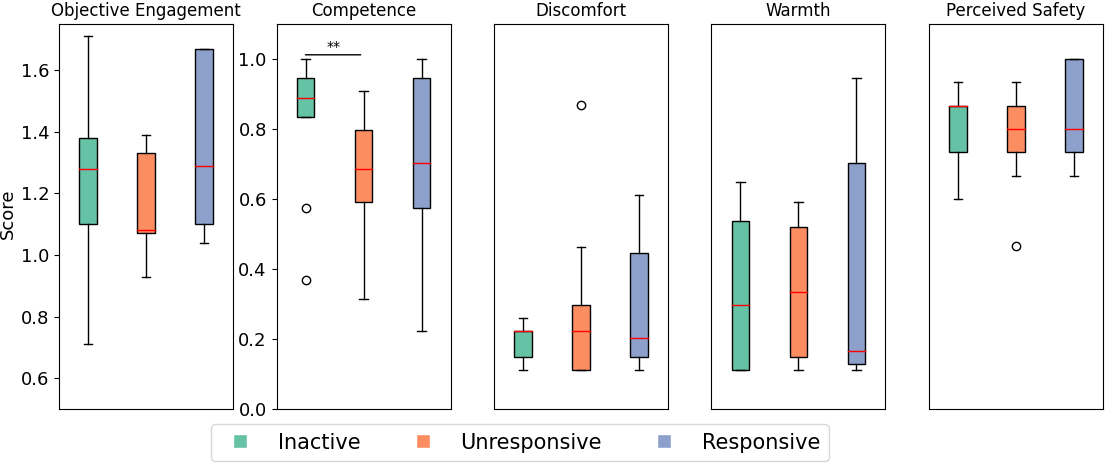}
    \caption{Boxplots illustrate participants' perceived competence, discomfort, warmth, safety, and objective engagement. Each box plot features a line representing the median, with box edges indicating the 25th and 75th percentiles. Whiskers display the range up to 1.5 times the interquartile range, with outliers marked as 'o'.  The ratio \(E_{obj}\)  is represented as a value potentially exceeding 1. Competence, perceived safety, discomfort, warmth, and safety ratings are normalized to fall within the range of 0 to 1. Significance levels ($\ast\ast := p-value < 0.01$) is indicated on lines between conditions.}
    \label{fig:ObjEngagement_Competence_Discomfort_Warmth_Safety}
\end{figure}
\begin{figure}[!h]
     \centering
     \begin{subfigure}[b]{0.49\textwidth}
         \centering
         \includegraphics[width=\textwidth]{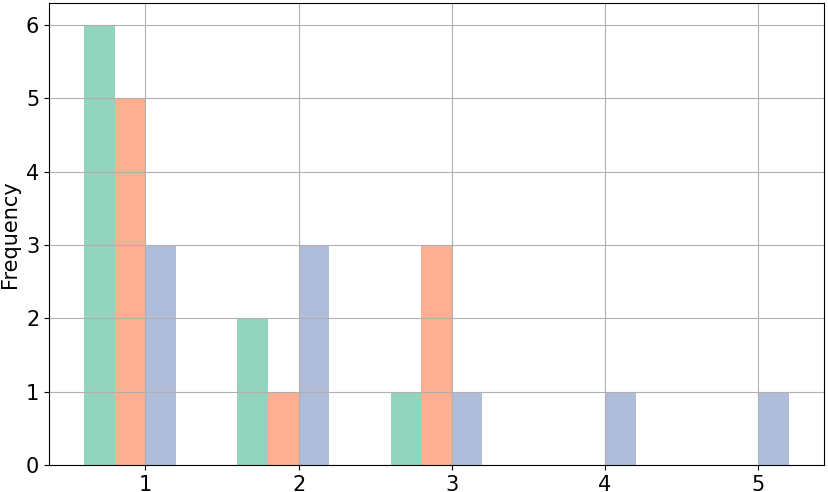}
         \caption{Trust}
         \label{fig_sub:trust}
     \end{subfigure}
     \hfill
     \begin{subfigure}[b]{0.49\textwidth}
         \centering
         \includegraphics[width=\textwidth]{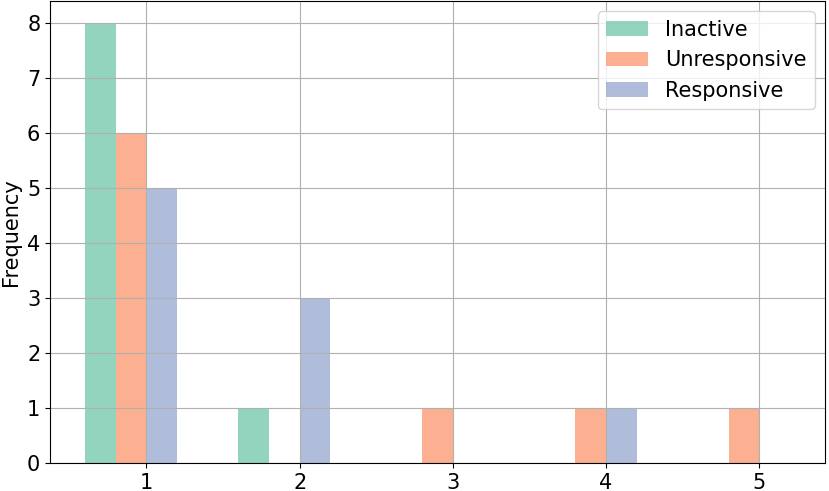}
         \caption{Subjective Engagement}
                  \label{fig_sub:subj_engagement}
     \end{subfigure}
        \caption{Bar plots illustrating participants' perceived trust and engagement for each condition on a 5-point Likert Scale (1 = highest, 5 = lowest).}
        \label{fig:trust_subj_engag}
\end{figure}

\section{Results}
\label{results}

% Descriptive Statistics Data Results
    % Robot Perception
To evaluate participants' perceptions of the robot, we calculated scores for competence, warmth, discomfort, and perceived safety by averaging individual items within each sub-scale. Figure \ref{fig:ObjEngagement_Competence_Discomfort_Warmth_Safety} illustrates participants' responses for each dimension across conditions. Notably, participants perceived the robot as more competent in the inactive condition compared to the active condition, showing relatively high ratings across conditions. Similarly, participants reported higher safety levels in the inactive condition compared to the active conditions, also displaying relatively high ratings overall. Ratings of discomfort remained consistently low across all conditions. Warmth ratings were comparable between the inactive and unresponsive conditions but were higher than the responsive condition, with relatively low ratings overall. Additionally, participants consistently expressed high levels of trust across all three conditions, as demonstrated in Figure \ref{fig_sub:trust}.
 
    % Engagement
Concerning engagement, participants consistently reported high levels across all three conditions, as shown in Figure \ref{fig_sub:subj_engagement}. During the actual session, participants executed a higher number of pushes, in accordance with the instructions, compared to the expected count calculated from the trial session. The inactive and responsive conditions showed nearly equal ratios, while the unresponsive condition had lower values in comparison to the other two conditions. Objective engagement is presented in the leftmost section of Figure \ref{fig:ObjEngagement_Competence_Discomfort_Warmth_Safety}. Objective and subjective engagement are presented in two separate graphs, as the former is continuous data and the latter is ordinal data. 

    % Inferential Statistical Analysis Procedure Only
    To analyze significant differences between the three facial expression conditions, we employed Mann-Whitney U and Chi-Square (\(\chi^2\)) tests. These non-parametric tests were chosen due to the smaller sample size in each condition \((n < 30)\), which precluded a normality check. The selection of tests was based on the type of data being compared – continuous and ordinal. To mitigate false positives, we applied the Bonferroni Correction test. Effect size calculations were performed for significant differences.
    
% Inferential Statistical Analysis
    % Robot Perception and Engagement Significant Difference Results
There was a statistically significant difference in competence scores between the inactive condition ($\mu = 8.24, \sigma = 0.55$) and the unresponsive condition ($\mu = 6.07, \sigma = 1.50$), with a \(p-value = 0.005\) and \(W = 58\). This difference was associated with a substantial effect size of \(r = 0.92\). No significant differences were observed for any other dimensions.

% Conclusion of Inferential Statistical Analysis - Answers to RQs
In conclusion, our findings in response to \textbf{RQ1} indicate that facial emotional expressions have a negative impact on older adults' perception of a robot, particularly in terms of competence. However, these expressions do not influence their level of engagement. As for \textbf{RQ2}, responsive robot facial emotional expressions, in comparison with unresponsive facial emotional expressions, do not have an impact on robot perception or engagement levels.

\begin{figure}[!htb]
     \centering
     \begin{subfigure}[b]{0.49\textwidth}
         \centering
         \includegraphics[width=\textwidth]{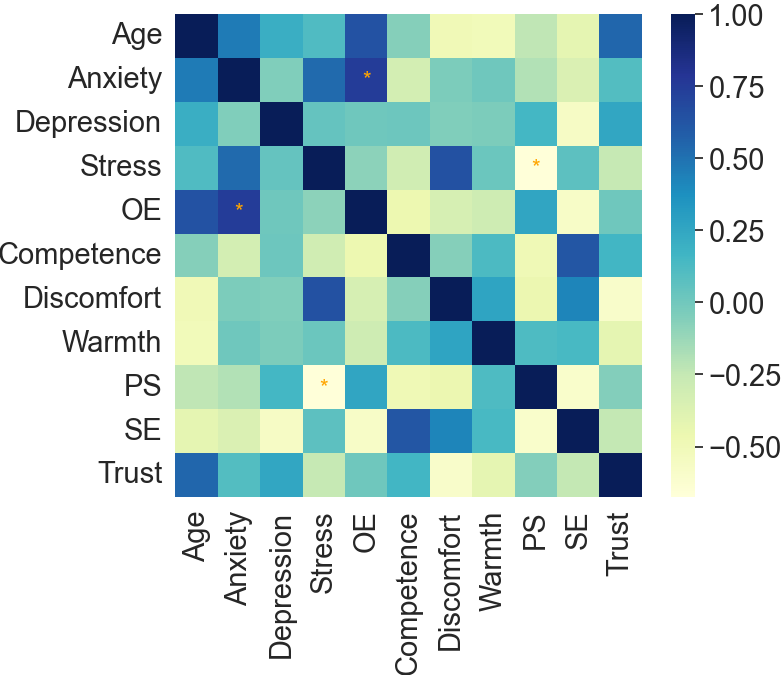}
         \caption{Inactive Correlations}
         \label{fig_sub:inactive_corr}
     \end{subfigure}
     \hfill
          \begin{subfigure}[b]{0.49\textwidth}
         \centering
         \includegraphics[width=\textwidth]{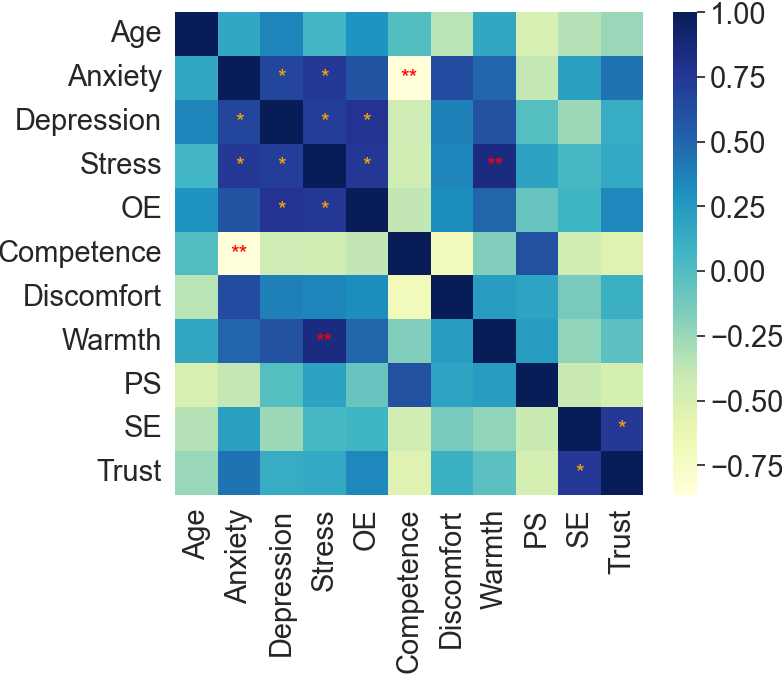}
         \caption{Unresponsive Correlations}
         \label{fig_sub:unresp_corr}
     \end{subfigure}
     \hfill
     \begin{subfigure}[b]{0.49\textwidth}
         \centering
         \includegraphics[width=\textwidth]{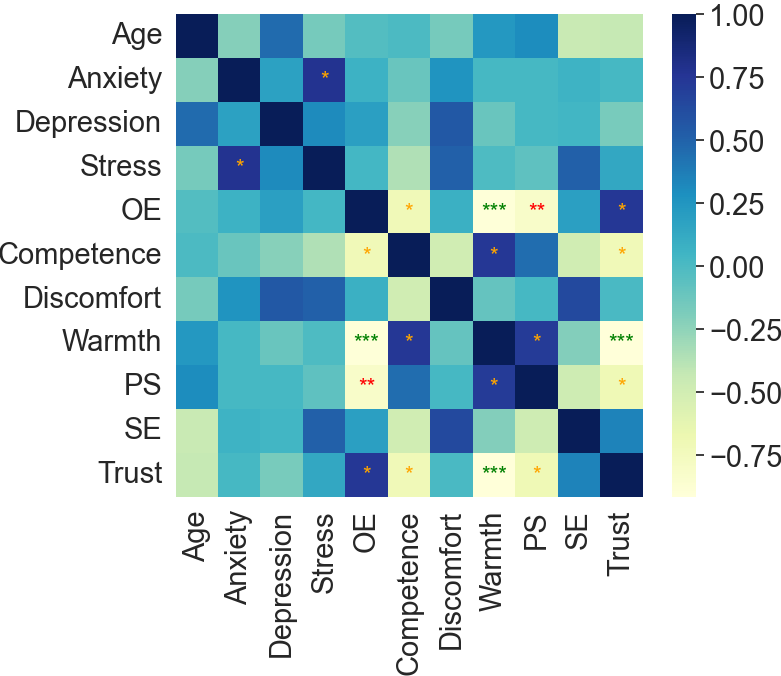}
         \caption{Responsive Correlations}
                  \label{fig_sub:resp_corr}
     \end{subfigure}
        \caption{Heatmap illustrates Spearman's correlations observed for each condition. Correlations marked with '*', '**', and '***' indicate a \(p-value < 0.05, < 0.01,\) and \(< 0.001\) with a \(\mid\rho\mid > 0.6\). Abbreviations: OE = Objective Engagement, SE = Subjective Engagement, PS = Perceived Safety.}
        \label{fig:correlations}
\end{figure}

\subsection{Correlation Analysis}
% Correlation Analysis Procedure Only
To enhance our comprehension of the quantitative results, we employed Spearman's rank correlation coefficient (\(\rho\)) to identify correlations within the collected data. For a deeper understanding, we investigated whether gender exerts an influence on any of the dependent variables. To achieve this, we calculated the Rank- Biserial correlation coefficient ($r_{rb}$) and conducted Mann-Whitney U and \(\chi^2\) tests. However, no correlations or significant differences were found across conditions with respect to gender.

% Correlations for Condition 1
        % Engagement Objective Measure	Anxiety	0.745122225	0.033884436
        % Perceived Safety	Stress	-0.674885239	0.046110007
In the inactive condition, as depicted in Figure \ref{fig_sub:inactive_corr}, it's apparent that individuals with lower stress levels tend to perceive the robot as safer, unlike those with higher stress levels. Conversely, increased engagement with the robot is associated with higher levels of anxiety.
    
% Correlations for Condition 2
    % Engagement Objective Measure	Depression	0.771186441	0.014960432
    % Engagement Objective Measure	Stress	0.750111499	0.019913383
    % Engagement Subjective Measure	Trust	0.741702467	0.022159077

    % Very Strong Correlations (rho > 0.8) with very likely significant differences (p-value < 0.005)
        % Competence	Anxiety	-0.864530967	0.002632372
        % Warmth	Stress	0.845688304	0.004072994
    Contrasting the inactive condition, the unresponsive condition reveals intriguing insights, characterized by a subset of robust correlations \((\mid\rho\mid > 0.8)\) with an exceedingly low likelihood of arising by chance \((p-values < 0.005)\). Illustrated in Figure \ref{fig_sub:unresp_corr}, participants exhibiting heightened anxiety levels are inclined to perceive the robot as less competent, contrasting those with lower anxiety levels. Additionally, individuals experiencing high stress levels are more inclined to perceive the robot as friendly and warm.

% Corrlations for Condition 3
    % Engagement Objective Measure	Competence	-0.716797615	0.029774043
    % Engagement Objective Measure	Perceived Safety	-0.812234819	0.007818601
    % Engagement Objective Measure	Trust	0.740152746	0.022590386
    % Competence	Warmth	0.745762712	0.021054935
    % Competence	Trust	-0.713286226	0.030968453
    % Warmth	Perceived Safety	0.7199346	0.02873281
    % Perceived Safety	Trust	-0.699142421	0.036095042

     % Very Strong Correlations with very likely significant differences
% Engagement Objective Measure	Warmth	-0.90136227	0.000900093
% Warmth	Trust	-0.913354313	0.000578746

In contrast to the previously mentioned conditions, the responsive condition exhibited robust correlations\((\mid\rho\mid > 0.8)\) with dependent variables, demonstrating a high degree of statistical significance \((p-values < 0.001)\). Participants who engage more intensively with the robot tend to perceive it as less warm than those who engage less. Similarly, participants reporting elevated levels of trust perceive the robot as less warm.

% Singificant correlations between RoSAS dimensions concerns us as Yue's paper and Hopko's paper said about Godspeed. 

% These very low significant differences in the correlations suggest that may be with a bigger sample size, we can find significant differences in Warmth 

It's important to keep in mind that anxiety, stress, and depression levels are assessed prior to the experiment. This means that the measurements reflect participants' baseline levels in their everyday lives rather than during the course of the experiment itself. Furthermore, no significant differences in anxiety, stress, and depression levels were found between the conditions.

\section{Discussion}
\label{discussion}

% Small sample size and its impact on results
We acknowledge that a larger sample size could have potentially revealed more statistically significant differences between dimensions, reducing the risk of Type II errors. However, due to the unique and challenging nature of recruiting older adults, obtaining a larger sample size was unfeasible. \modtext{Our sample size is greater than previous studies involving robots and older adults, such as Nowak et al. \cite{nowak_assistance_2022}, who could only recruit 7 participants, and Giorgi et al. \cite{giorgi_friendly_2022}, who recruited 17 participants to investigate perceived trust in a social robot across a wider age range (40 to 87 years).} Moreover, the study's design, conducted as a between-subject experiment to mitigate issues highlighted in prior literature \cite{fitter_how_2020}, unintentionally accentuated differences between conditions, reinforcing the need for a greater participant count.

% Significant Difference in Robot Perception between Inactive and Unresponsive
Our findings demonstrate that participants perceived the robot as less competent when facial expressions were introduced, as opposed to when they were absent. This observation can be attributed to the non-social nature of the pHRI application—physical exercise—employed in our experiment. In contrast, previous literature has shown that incorporating facial expressions into pHRI applications with a social purpose, such as handshakes, led participants to perceive the robot's emotions more positively, as evidenced by Tsalamlal et al. (2015) \cite{tsalamlal_affective_2015}. Furthermore, this observation is supported by participants' responses to our open-ended questions, in which they referred to the robot as a "tool" or a "weight machine" in both active and inactive conditions. This underscores the importance of future investigations to enhance the robustness of these findings across various pHRI applications.

Moreover, it's worth considering the distinction between the physical interaction described in Tsalamlal et al.'s study \cite{tsalamlal_affective_2015}, where a humanoid was employed, and our utilization of a collaborative robot like Sawyer, equipped with a screen. This contrast prompts a pertinent inquiry regarding how the type of robot employed in pHRI might influence robot perception. It's conceivable that people's preferences for engaging in physical interaction and communication could differ when interacting with a humanoid displaying facial expressions compared to a collaborative robotic arm with a screen conveying facial expressions. \modtext{However, we believe that the context, whether social or collaborative, significantly influences user communication preferences.}

Additionally, a notable recommendation arises to enhance the robot's facial features by incorporating elements such as cheeks or eyelids, aligned with the findings from Chen and Jia's study \cite{chen_effects_2023}, thereby imparting a sense of maturity and user preference to the robot's appearance. This recommendation finds additional support through a considerable number of responses obtained from participants in our study's open-ended questions, wherein many participants suggested altering the robot's facial features. One participant even specifically mentioned the eyes. Implementing this adjustment holds the potential to cultivate a perception of Sawyer as more mature and visually appealing.

% Significant Difference in Robot Perception between Inactive and Responsive
We anticipate that a comparison between the inactive and responsive conditions would likely unveil a significant difference in competence, given the availability of a larger sample size. Importantly, this observation extends to all other dimensions, underlining the pivotal role of conducting the study with a more substantial number of participants. Achieving this can involve augmenting the relatively hard-to-access population of older adults with a younger adult population. This augmentation naturally leads to the question of whether distinct age groups will perceive the robot's attributes in varying ways. \modtext{ Therefore, we plan to expand this research in the future by including  younger adults, following the common practice in robotics studies involving older adults \cite{nowak_assistance_2022}.}

% Significant Difference in Robot Perception between Unresponsive and Responsive
We speculate that the absence of a significant difference between the responsive and unresponsive conditions might be attributed to the abstract nature of the chosen set of facial expressions and the specific type of application employed in this experiment. This conjecture is supported by the fact that more than 50\% of participants in the responsive condition reported not perceiving any changes in the robot's facial expressions, as their attention was directed towards activating the headlight blinks. This perspective gains further weight from one participant's response to the open-ended questions, where they suggested altering the facial expressions once the exercise was completed. This observation underscores the possibility that the nature of the application and its objective could have influenced participants' focus on facial expressions. It further implies that pHRI applications primarily centered around collaborative tasks, as opposed to sociability, such as engaging in physical exercises instead of handshaking, should consider incorporating facial features that are less abstract and more conspicuous.

% Significant Difference in Engagement between Inactive, Responsive, and Unresponsive
The absence of a significant difference in participants' levels of engagement across all conditions suggests that facial expressions may not exert any influence on engagement during specific types of pHRI applications. However, an argument can be made that the robot's joint impedance control inhibited certain participants from executing a high number of pushes, as they waited for the robot's arm to return to its initial position after each push. In contrast, other participants proceeded without such hesitations. This divergence in exercise approaches highlights a potential limitation: the count of pushes might not have been the most suitable metric for assessing objective engagement in this context. \modtext{Therefore, we encourage researchers to explore better-suited engagement measures.}

Similarly, the lack of statistically significant differences in the subjective engagement measure bolsters the notion that facial expressions might not significantly impact participants' levels of engagement. This interpretation gains further support from the notable proportion of participants who recommended additional features like rhythmic music, encouraging phrases, visual timers, and verbal motivation. Collectively, these inputs suggest that facial expressions might not be the most suitable non-verbal social cue for enhancing engagement in collaborative pHRI applications.

% Correlation Analysis
There were no identifiable correlations between the conditions (each condition displayed different correlations), therefore the direct relationship between the correlations and the conditions remains uncertain. Nevertheless, the correlation analysis suggests that people's perceptions of a robot's attributes (safety, competence, friendliness, warmth) can be influenced by their anxiety, stress levels, engagement with the robot, and trust, and this could vary depending on the experimental condition. 
Notably, anxiety and stress levels significantly influence these perceptions. People with lower stress felt the robot was safer, while those with heightened anxiety perceived it as less competent.
Engagement levels played a key role too. In the inactive condition, deeper engagement raised anxiety, while in the responsive condition, intense engagement led to a perception of reduced warmth. Trust further added complexity, as higher trust levels correlated with perceiving the robot as less warm, reflecting a more critical assessment.

These findings underline that emotions are not peripheral to pHRI; they are integral. However, a clear relationship between these findings and the distinct experimental conditions has yet to be established. To gain a more comprehensive understanding, additional research studies are necessary but out of the scope of the current paper.

\section{Conclusion}
\label{conclusion}

Our research aimed to uncover the impact of facial expressions on a robot's perceived attributes and an individual's level of engagement within a collaborative pHRI application, with a particular focus on older adults. Our findings demonstrated that when robots display facial expressions during a collaborative pHRI application, people tend to perceive them as less intelligent compared to robots that do not exhibit any facial expressions. Interestingly, we observed that these facial expressions do not significantly influence the levels of engagement among older adults.
We speculate that participants' perception of robots is intricately tied to the collaborative nature of pHRI applications we emphasized in our study, as opposed to socially oriented pHRI applications. Our results suggest the need for more appropriate non-verbal social behaviors to enhance participants' engagement levels.
In the next experiments, we will also consider a larger number of participants by including younger adults, which will also allow us to investigate possible differences due to age groups, and further investigate the relationships uncovered by the correlation analysis with respect to the different conditions.

\printbibliography

\end{document}